\documentclass[aps,twocolumn,superscriptaddress]{revtex4-1}
\usepackage{amsmath, amsfonts, graphicx}
\usepackage{subfigure}
\usepackage[citecolor=blue]{hyperref}
\usepackage{bm}
\usepackage[framemethod=tikz]{mdframed}
\usepackage{xpatch}
\usepackage{makecell}
\usepackage{url}
\usepackage{float}
\usepackage[ruled]{algorithm2e}

\newcommand{\ii}{\mathrm{i}}
\newcommand{\ee}{\mathrm{e}}

\def\be{\begin{equation}}
\def\ee{\end{equation}}
\def\bee{\begin{eqnarray}}
\def\ene{\end{eqnarray}}
\def\bes{\begin{subequations}}
	\def\ees{\end{subequations}}

\begin{document}

	\title{Measuring the rogue wave pattern triggered from Gaussian perturbations by deep learning}
	
	\author{Liwen Zou}
	
	\author{XinHang Luo}
	
	\author{Delu Zeng}
	\email{Corresponding author: dlzeng@scut.edu.cn}
	
	\author{Liming Ling}

	\affiliation{School of Mathematics, South China University of Technology, Guangzhou 510640, China}
	
	\author{Li-Chen Zhao}
	
	\affiliation{School of Physics, Northwest University, Xi'an 710127, China}
	
	
	\begin{abstract}
		Weak Gaussian perturbations on a plane wave background could trigger lots of rogue waves, due to modulational instability. Numerical simulations showed that these rogue waves seemed to have similar unit structure. However, to the best of our knowledge, there is no relative result to prove that these rogue waves have the similar patterns for different perturbations, partly due to that it is hard to measure the rogue wave pattern automatically. In this work, we address these problems from the perspective of computer vision via using deep neural networks. We propose a Rogue Wave Detection Network (RWD-Net) model to automatically and accurately detect RWs in the images, which directly indicates they have the similar computer vision patterns. For this purpose, we herein meanwhile have designed the related dataset, termed as Rogue Wave Dataset-$10$K (RWD-$10$K), which has $10,191$ RW images with bounding box annotations for each RW unit. In our detection experiments, we get $99.29\%$ average precision on the test splits of the RWD-$10$K dataset. Finally, we derive our novel metric, the density of RW units (DRW), to characterize the evolution of Gaussian perturbations and obtain the statistical results on them.
		
	\end{abstract}
	
	
	

	\maketitle

	\section{Introduction}
	The focusing nonlinear Schr\"odinger equation (NLSE)
	\begin{equation}\label{eq:nls}
	\ii {u}_t+\frac{1}{2}{u}_{xx}+|{u}|^2{u}=0
	\end{equation}
	is a universal model in the nonlinear science such as the deep water wave \cite{RW_tank}, nonlinear optics \cite{prw,rw2010,Dudley2014}, Bose-Einstein condensate \cite{bec-rw} and even the finance \cite{yanfrw}. The rogue wave (RW) solution or the Peregrine soliton is a typical exact solution for this model, which is related with the modulational instability (MI) \cite{Kharif2009,Onorato}. Even though the NLSE is an integrable model and possesses lot of exact solutions (solitons, multi-solitons, breathers, rogue waves), which had been observed in the physical experiments. Lots of rogue wave patterns had been discovered and analyzed based on analytic solutions  \cite{Kedziora2013,YangY-21}.  However, it is crucial to consider the general initial data problem for the NLSE, since many different waves can be generated by MI on a plane wave background.

	The optical rogue waves are observed in the integrable turbulence for the NLSE \cite{Walczak-15}, which is related to random initial data. Recent studies also show that RW patterns and integrable turbulence are observed in the soliton gas, for which the statistics on the kinetic, potential energies and other features are performed to characterize  integrable turbulence \cite{Gelash-18-19}. Starting from the stochastic perturbation on the non-zero background, the maximum peaks and probability density functions on the intensity are analyzed by the numerical method \cite{Soto-Crespo-16}, which provides the explanations on the probability of the appearance of rogue waves in a chaotic wave state. In the previous studies, the turbulence is measured by the statistics. If we depart from arbitrary initial data on the plane wave background, the chaotic structures can be observed from the numeric simulation due to the development of MI \cite{MI-1967,Baronio,MI-nonlinear-MI}, in which the weak units similar as the Peregrine solitons can be detected clearly. Thus, it is naturally to measure the RW pattern and compute the density of RW by the RWs number. However, the bijection relation between RWs and the scattering data has not been established, in contrast to soliton density in soliton gas, which is well established by inverse scattering theory \cite{Gelash-18-19}.
	
	The machine learning method \cite{Jordan-15} was used to analyze the extreme events in optical fiber MI \cite{Narhi-18}, in which the intensity and spectral intensity are analyzed by the supervised and unsupervised learning method. The machine learning algorithms, which involve the $k$-nearest neighbors, support vector machine and artificial neural networks, were also used for predicting the amplitude in the chaotic lase pulse \cite{Amil-19}. The extreme events are predicted by the deep neural networks in a truncated Korteweg-de Vries (tKdV) statistical framework \cite{Qi-20}. Other relative statistic learning methods to the extreme events are performed by the literature \cite{Mohamad-18,Dematteis-18,Majda-19,Salmela-20}. Object detection based on deep learning is widely applied to various scientific fields, e.g. Tao et al. combined microscopy observation with artificial neural networks (ANNs) and realized by machine learning algorithms for the study of starch gelatinization \cite{Tao}.  A numerical method based on the neural network can be used to long time simulation of rogue waves and other nonlinear waves having the MI \cite{WangLZF-21}. Recent developed machine learning method provides possibilities to measure the RW pattern and compute the density of RWs.
	
	In this work, we propose our deep neural networks (DNNs) model to automatically and accurately detect the RWs in the images. We have designed the related dataset, which has $10,191$ RW images with bounding box annotations for each RW unit. We further derive our novel metric, the DRW, to characterize the evolution of Gaussian perturbations, and finally give the statistical characters on them.
	
	\section{Preliminary}
	
\begin{figure}[h]
		\centering
		{
			\includegraphics[width=8.5cm,]{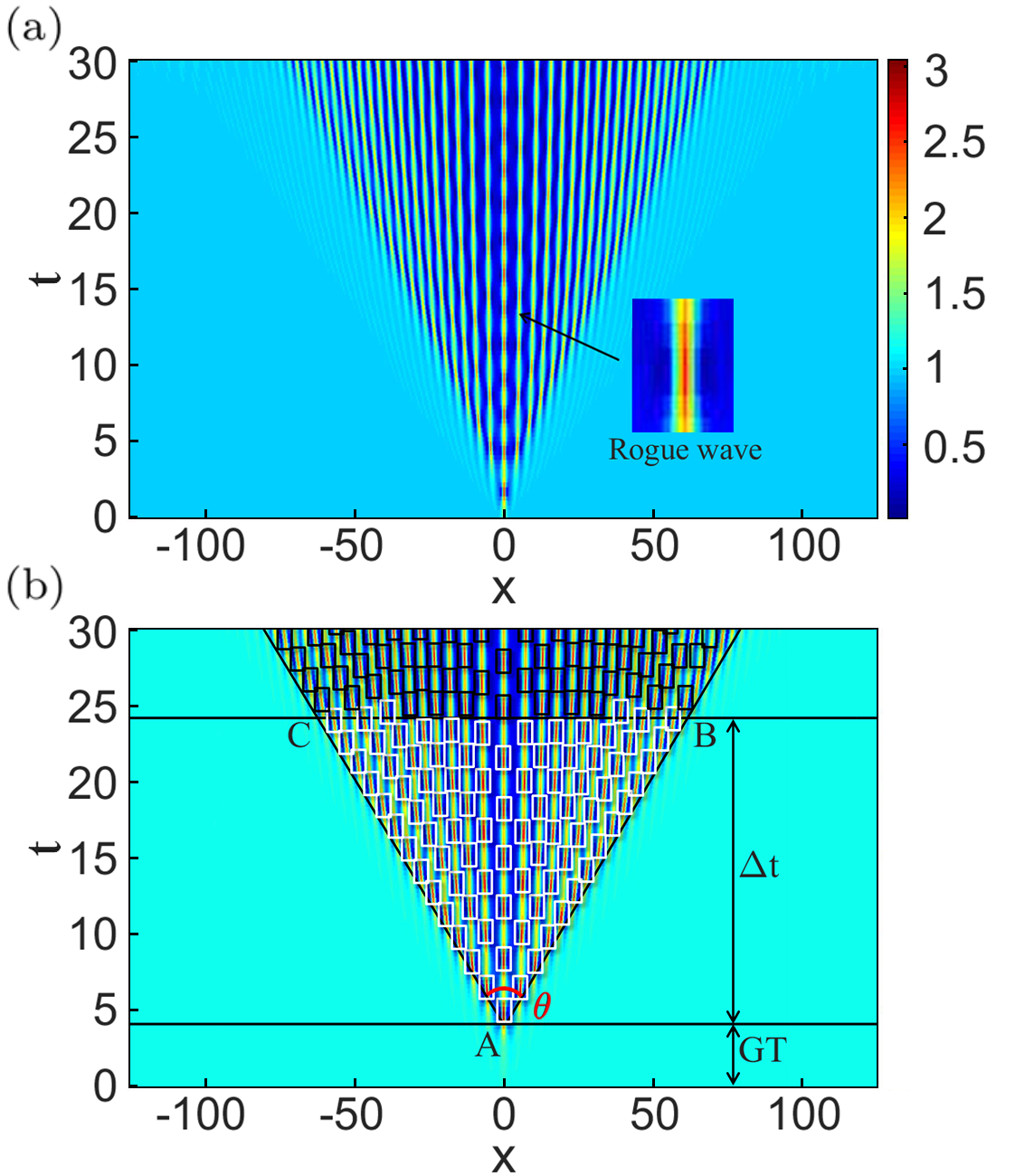}
		}
		\caption{(a) Rouge waves on rogue wave pattern images; (b) Some terms defined in this paper.}\label{fig1}
	\end{figure}

	It has been shown that local weak perturbations can generate RWs \cite{Zhaoling,GaoZYLY-20}. We therefore consider an  initial data $u(x,0)=1+p(x)$ (where $p(x)$ is a white noise or weak localized perturbation) to investigate the evolution patterns containing RWs.
	By using the $4$th-order integrating-factor method for solving the NLSE \cite{Yang-10}, we see that the initial data
	can generate lots of RWs with time evolution. Interestingly, each of them has the similar shape as the fundamental RW (Peregrine breather), which has the form \cite{Peregrine1983,Akhmediev2009a}:
	\begin{equation}\label{eq:rw}
	u_{\rm P}(x,t)=\left[1-\frac{4(1+2\ii t)}{1+4(x^2+t^2)}\right]{\rm e}^{\ii t}.
	\end{equation}
	whose temporal-spatial pattern is shown in FIG. \hyperref[fig1]{1(a)}.   However, there is no systematic evidence to illustrate the results, which is \textbf{an open problem} for us up to now.   In this paper, we would like to solve this problem by the artificial neural network. We choose an initial data with usual Gaussian form perturbations \cite{Zhaoling}
	\begin{equation}\label{eq:initial}
	u(x,0)=1+2\varepsilon^{-1}{\rm e}^{-\frac{x^2}{\mu^2}},
	\end{equation}
	to demonstrate our results. The corresponding Lax spectrum of the initial data was found to locate at the pure imaginary axis \cite{Biondini-18}. The high-order RWs can be generated by the multi-Gaussian perturbations on a plane wave background \cite{GaoZYLY-20}. To measure the RW pattern evolved from the Gaussian perturbation, we define the density of rogue wave (DRW) as follows:
	\begin{equation}\label{eq:DRW}
	{\rm DRW}=\frac{N(\Delta t,\varepsilon,\mu)}{S_{\Delta ABC}}=\frac{N(\Delta t,\varepsilon,\mu){\rm cot}\frac{\theta(\varepsilon,\mu)}{2}}{(\Delta t)^2}
	\end{equation}
	where \(N(\Delta t,\varepsilon,\mu)\) denotes the number of RWs that appear during the period from \(t=GT\) to \(t=GT+\Delta t\).  \textbf{We compute $N$ according to the number of RW bounding boxes in \(\Delta ABC\) and incomplete boxes are computed by the proportion of their areas inside \(\Delta ABC\), for instance, half of a box is computed as $0.5$}. \(S_{\Delta ABC}\) is the area of the triangular \(\Delta ABC\). It should be pointed out that \(GT\) is the time interval between the beginning moment and the time when the first RW peak appears. Fig. \hyperref[fig1]{1(b)} shows the details of some terms defined in this paper. Nextly, we introduce our DNNs model to detect RW pattern.
	
	
	\section{RWD-Net}
	Image recognition and object detection by deep learning methods had gotten great development in recent years.  Residual Network (ResNet) was designed to learn the residual representation between the layers, which not only train the network easier but also improve the accuracy a lot \cite{He2016}. Even now ResNet is still one of the most fundamental feature extraction network in the computer vision field.  Lin et al. proposed the Feature Pyramid Network (FPN) which improves the accuracy of object detection, especially in the detection of weak objects, without adding additional calculations by extracting and fusing multi-scale feature map of RW images \cite{He2017b}. Based on the ResNet and FPN, Lin et al. proposed their RetinaNet with the noval focal loss function replacing the traditional cross-entropy loss to solve the unbalance problem between positive and negative objects, which improves both the accuracy and speed \cite{He2017a}. Goldman et al. improve the detection in densely packed scenes by adding the Soft-IoU layer and the Expectation-Maximization Merge (EM-Merge) unit to the RetinaNet \cite{Goldman2019}.
	
\begin{figure}[h]
		\centering
		{
			\includegraphics[height=4.5cm]{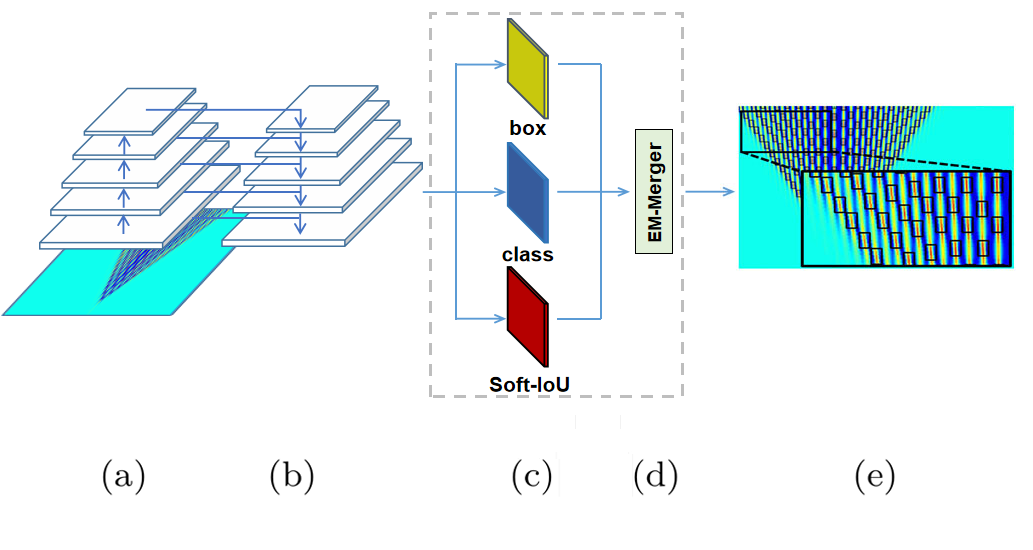}
		}
		\caption{Structure of our RWD-Net. (a) ResNet-101; (b) FPN; (c) The box, class and the Soft-IoU subnets; (d) EM-Merge unit; (e) Detection results. }\label{fig2}
	\end{figure}

	Inspired by the studies above, we propose a new methodology, namely Rogue Wave Detection Network (RWD-Net), developed from RetinaNet improved by Goldman et al. in \cite{Goldman2019}, which is designed to detect the RW regions in the images and get the number and distribution of them to measure the rogue wave pattern. The architecture of the network is shown in Fig.\ref{fig2}, which mainly consists of three parts: feature extractor or backbone, detector and post-processing unit. The backbone is composed of ResNet and FPN to get $5$ feature maps of RW images under different scales. Each instance in the feature map is covered with $9$ priori boxes, called anchors, each of which has different sizes and shapes and will be send to the next $3$ subnets---class, box and Soft-IoU subnet to be trained.

	Mathematically, let the pixel values of a RW image be denoted by \(I\in\mathbb{R}^{W\times H \times L}\), \(W,H,L\in \mathbb{Z}^+\) denotes the width, height and length of the image respectively, and the training data is denoted by \(\mathcal{D}^{tr}=\left\{(I^l,Y^l)\right\}_{l=1}^L\), where \(Y^l\in\mathbb{R}^{5\times N}\) is the set of offset and label \(\left(d_a^x,d_a^y,d_a^w,d_a^h,y_a\right)\) of each anchor covering \(I^l\) and \(N\) denoted the number of anchors in image \(I^l\). Our goal is to predict the bounding box matrices \(\hat{Y}^l\in\mathbb{R}^{5\times N}\) of image \(I^l\). A RW image \emph{I} is first fed into the backbone consists of ResNet and Feature Pyramnd Network (FPN), parameterized by \(\Theta\), to produce a feature map \(F(I;\Theta)\in\mathbb{R}^{W'\times H'\times L'\times C}\), where \emph{C} is the number of feature channels. The feature vector \(f_a(I;\Theta)\in\mathbb{R}^C\) at each spatial location $a$ on the feature map \(F(I;\Theta)\) represents the extracted feature of anchor $a$ on the image \emph{I} by the ResNet+FPN model. After the feature extracting, there are $3$ subnets of our proposed RWD-Net: the class subnet for RW/background classification of each anchor feature \(f_a(I;\Theta)\), the box subnet for the bounding box regression and the Soft-IoU subnet for predicting the Soft-IoU between the prediction boxes and the ground truth.
	
	In the class subnet, we define the confidence level that an anchor instance \(f_a(I;\Theta)\) covers a RW region as:
	\begin{equation}\label{}
	p_a(I;\Theta,w_c)=\frac{1}{1+\exp(-w_c^T f_a(I;\Theta))}
	\end{equation}
	where \(w_c\in\mathbb{R}^C\) is the parameter of the class subnet, then we minimize the focal loss on the traing data with annotations:
	\begin{equation}\label{}
	\begin{split}
	&l_{class}(I^l,Y^l;\Theta,w_c)\\
	=&-\sum_{a\in A}[y_a^l \alpha(1-p_a(I;\Theta,w_c)^\gamma )\log p_a(I;\Theta,w_c)\\
	&+(1-y_a^l)(1-\alpha)p_a(I;\Theta,w_c)^\gamma \log(1-p_a(I;\Theta,w_c))]
	\end{split}
	\end{equation}
	where \(y_a^l\) is the label annotation of anchor $a$ obtained directly from \(Y^l\) and $A$ is the set of all anchors covering the image \(I^l\). We set \(\alpha=0.25\) and \(\gamma=2\) in our implementation. And the loss function for training the class subnet over the whole traing set \(D^{tr}\) is defined by
	\begin{equation}\label{}
	\begin{split}
	&L_{class}(D^{tr};\Theta,w_c)\\
	=&\frac{1}{\big|D^{tr}\big|}\Sigma_{(Z^l,Y^l)\in D^{tr}}l_{class}(I^l,Y^l;\Theta,w_c).
	\end{split}
	\end{equation}
	
	We now introduce the box subnet module of our RWD-Net. The model predicts the offset between positive anchors and their belonged ground truth as
	\begin{equation}\label{}
	\hat{d}_a=(\hat{d}_a^x,\hat{d}_a^y,\hat{d}_a^w,\hat{d}_a^h)=w_b^Tf_a(I;\Theta)
	\end{equation}
	where \(w_b\in\mathbb{R}^{C\times 4}\) is the parameter of box subnet. Then we use the smooth \(-L_1\) loss function \cite{smoothL1} on the image \(I^l\) with box annotations to learn the box subnet:
	\begin{equation}\label{}
	l_{box}(I;\Theta,w_b)=\Sigma_{a\in A}y_aQ_a
	\end{equation}
	where
	\begin{equation}\label{}
	Q_a=
	\left\{
	\begin{array}{lr}
	\frac{1}{2}\left\|\hat{d}_a-d_a\right\|_2^2, &if\,\left\|\hat{d}_a-d_a\right\|_2<1  \\
	\left\|\hat{d}_a-d_a\right\|_2-\frac{1}{2}, &otherwise \\
	\end{array}
	\right.
	\end{equation}
	and \(d_a=(d_a^x, d_a^y,d_a^w,d_a^h)\) is the annotation of anchors' offset. Now we can write down the loss function on the trainning data \(D^{tr}\):
	\begin{equation}\label{}
	L_{box}(D^{tr};\Theta,w_b)=\frac{1}{\big|D^{tr}\big|}\sum_{(Z,Y)\in D^{tr}}l_{box}(I;\Theta,w_b).
	\end{equation}
	
	The Intersection over Union (IoU) defined as
	\begin{equation}\label{}
	IoU=\frac{I(X)}{U(X)}
	\end{equation}
	is used as the evaluation metric in the detection, where \(I(X)\) and \(U(X)\) denote the intersection and union area between the prediction and ground truth respectively. The class subnet gives the confidence and the Soft-IoU subnet gives the Soft-IoU which is just the IoU predicted by the model and is not the real IoU. The Soft-IoU between the prediction boxes and their belonged ground truth is defined as
	\begin{equation}\label{}
	s_a(I;\Theta,w_s)=w_s^Tf_a(I;\Theta)
	\end{equation}
	where \(w_s\in\ \mathbb{R}^C\) is the parameter of the Soft-loU subnet. Then we minimize the cross-entropy loss function on the image \(I^l\):
	\begin{equation}\label{}
	\begin{split}
	&l_{s-iou}(I^{l},Y^{l};\Theta,w_s)\\
	=&-\sum_{a\in A}[IoU_a\log s_a(I;\Theta,w_s)\\
	&+(1-IoU_a)\log(1-s_a(I;\Theta,w_s))]
	\end{split}
	\end{equation}
	where \(IoU_a\) is the IoU ground truth between anchor in location \(a\) and their belonged ground truth. And the whole loss in this Soft-IoU subnet for training data \(D^{tr}\) defined as:
	\begin{equation}\label{}
	L_{s-iou}(D^{tr};\Theta,w_s)\\
	=\frac{1}{\big|D^{tr}\big|}\sum_{(Z^l,Y^l)\in\\ D^{tr}}l_{s-iou}(I^l,Y^l;\Theta,w_s)
	\end{equation}
	
	Finally, we write down the overall loss function for traing our RWD-Net:
	\begin{equation}\label{}
	\begin{split}
	&L_{RWD}(D^{tr};\Theta,w_c,w_b,w_s)\\
	=&L_{class}(D^{tr};\Theta,w_c)+L_{box}(D^{tr};\Theta,w_b)\\
	&+L_{s-iou}(D^{tr};\Theta,w_s)
	\end{split}
	\end{equation}
	All parameters are jointly optimized during network training. The optimized parameters are obtained by
	\begin{equation}\label{}
	\begin{split}
	&(\Theta^*,w_c^*,w_b^*,w_s^*)\\
	&=\underset{\Theta,w_c,w_b,w_s}{\arg\min}L_{RWD}(D^{tr};\Theta,w_c,w_b,w_s)
	\end{split}
	\end{equation}
	And we minimize the overall loss function by stochastic gradient decent method.
	
	
	Given a testing image \(I\), whether its anchor in location \(a\) covers RW regions is determined by:
	\begin{equation}\label{}
	\hat{y}_a=
	\left\{
	\begin{array}{lr}
	1, &if\; p_a(I;\Theta^*,w_c^*)\geq 0.5  \\
	0, &otherwise
	\end{array}
	\right.
	\end{equation}
	If \(\hat{y}_a=1\), then we further need to do the bounding box regression for anchor in \(a\):
	\begin{equation}
	\hat{d}_a=(w_b^*)^\mathrm{T}f_a(I;\Theta^*)
	\end{equation}
	
	With \(\hat{y}_a\) and \(\hat{d}_a\) above, we obtain the prediction \(\hat{Y}\). But it does not end, RWD-Net also predicts the Soft-IoU between each positive anchor and the ground truth:
	\begin{equation}
	\hat{s}_a=(w_s^*)^\mathrm{T}f_a(I;\Theta^*)
	\end{equation}
	which is used for expectation-maximization-merge (EM-Merge) unit next to obtain the final predicting boxes.  The details about the EM-Merge unit can be seen  in \cite{Goldman2019}.

	\section{Experimental results}
	In this work, we provide our big dataset, termed as Rogue Wave Dataset-$10$K (RWD-$10$K) containing 10,191 images of rogue wave pattern. We propose an efficient semi-automatic method, called Peak Search method, to achieve fast and accurate pre-detection to rogue waves instead of manual labeling. It is designed to determine the approximate location of the peak for each rogue wave by filtering out the local maximum points on the numerical solution matrices such that one peak point on the matrix corresponds to one rogue wave on the image. Then, all of these peak points will be expanded into bounding boxes. It should be noted that the sizes and locations of these bounding boxes will be manually refined due to the errors caused by the dimensional difference between the numerical solution matrices and the image matrices. By Peak Search and additional minor corrections, we can efficiently build the RWD-10K dataset. The details about Peak Search are given in the Appendix A.

	We variate the parameters \(\mu\) (from $0.5$ to $50$ with the interval $0.5$) and \(\varepsilon\) (from $11$ to $110$ with the interval $1$) in the initial NLSE \eqref{eq:nls} to generate $10000$ images. Additionally, to study the variation of DRW or other terms for \(\mu\) and \(\varepsilon\), we make other $191$ images with larger value range of \(\mu\) and \(\varepsilon\) to complete the following statistical experiments. Then we get the pseudo annotations by the Peak Search algorithm and refine manually so that we assembled our big benchmark RWD-10K. Each image corresponds a parameter two-tuple (\(\varepsilon\), \(\mu\)) in the initial equation. We focus on such settings for two reasons. First, every image in our dataset has its physical meaning. Second, by detecting those images we can capture the pattern similarity of computer vision among them and get certain statistical results about the distribution of these rogue waves (e.g. the evolution that the number of the rogue wave changes with the parameters in the initial data \eqref{eq:nls})
	
	In our detection experiment, the RWD-10K dataset is partitioned into train, validate and test splits. Training consists of 60\% of the images ($6071$ images) and their associated $510,431$ bounding boxes; $20\%$ of the images ($2024$ images), are used for validation (with their $172,881$ bounding boxes).  The rest $2024$ images  ($175,968$ bounding boxes) were used for testing. Images were selected randomly, ensuring that the same RW from the same image does not appear in more than one of these subsets. We present our RWD-10K dataset open for public studies and you can get more details at \href{https://github.com/ZouLiwen-1999/RogueWave}{https://github.com/ZouLiwen-1999/RogueWave}.
	
\begin{figure}[h]
		\centering
		{
			\includegraphics[width=8.6cm]{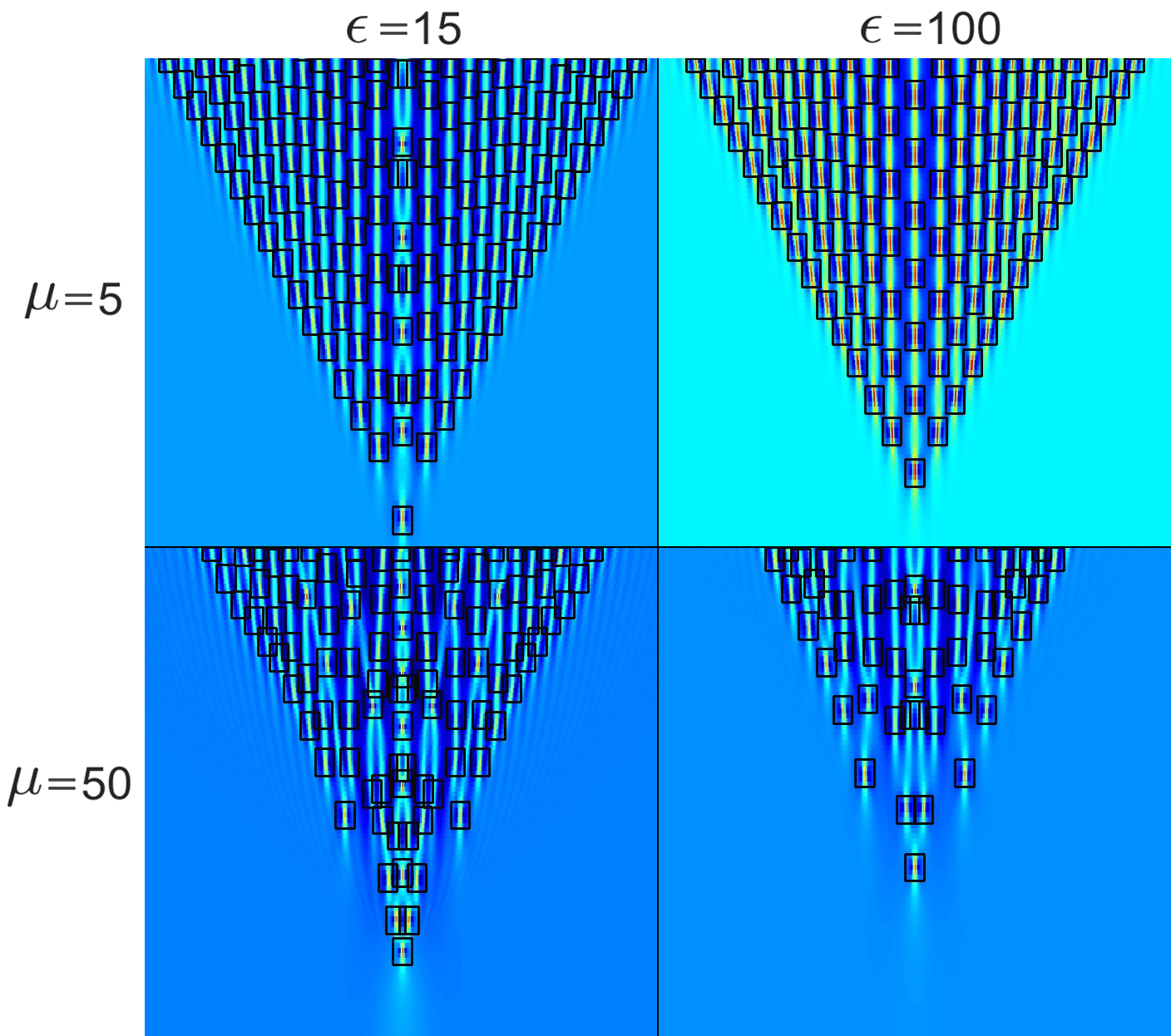}
		}
		
		\caption{The detection results of the images we randomly selected from the test splits of RWD-10K using our RWD-Net. }\label{fig3}
	\end{figure}

	The details about our experiment settings are shown in Appendix C. After 20 epochs training, we get the average precision (AP) of $99.29\%$ in the RW detection experiment, which shows that we successfully capture the rogue wave pattern similarity from the perspective of computer vision. Fig.\ref{fig3} shows the detection results of the images whose parameters are randomly chosen from the test dataset and we can see that we shot almost every rogue wave in the detection. Additionally, it is easy to observe that the distribution of these rogue wave patterns under different initial data are discriminative, which prompts us to do further statistics in the next section.

	\section{Measuring the rogue wave pattern}
\begin{figure}[h]
		\centering
		{
			\includegraphics[height=6.6cm]{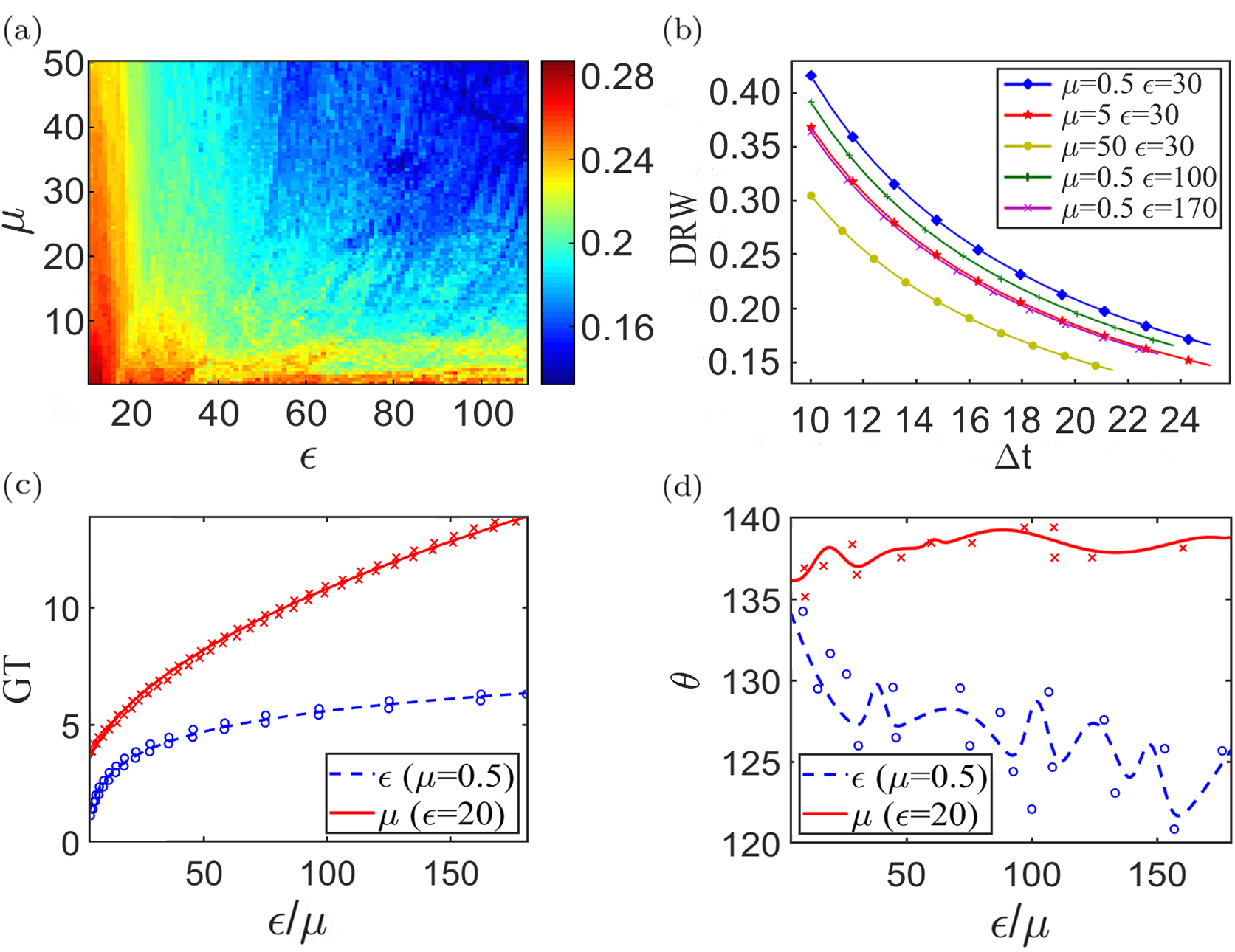}
		}	
		\caption{Results of measuring the rogue wave pattern. (a) The map of DRW with the change of \(\varepsilon\) and \(\mu\) when \(\Delta t=15\) ; (b) The variations of DRW for different \(\Delta t\) ; (c) The relation between GT and \(\varepsilon\) or \(\mu\), in which the solid and blue lines are the fitting curves and the cross or circle points are the data points; (d) The \(\theta\) values with different \(\varepsilon\) or \(\mu\) values. }\label{fig4}
	\end{figure}

	Now we try to measure the RW pattern under different Gaussian initial data by the trained RWD-Net. We use the DRW defined in \eqref{eq:DRW} to quantify the RW pattern and we can also get the variation of GT and \(\theta\) based on the detection results of our RWD-Net model. The results are shown in the Fig. \ref{fig4} and initial data is given in \eqref{eq:initial}.
	
	FIG. \hyperref[fig4]{4(a)} is the distribution of the DRW with respect to the parameter \(\varepsilon\) and \(\mu\)  when \(\Delta t\) in \eqref{eq:DRW} is fixed as 15.
	Meanwhile, it is easy to find that the value of DRW decreases from bottom left to top right of the image, which means the value of corresponding DRW will decline in the spatiotemporal regions when \(\varepsilon\) and \(\mu\) increase. These characters can be understood by a fact that each localized wave is closer to Peregrine RW for initial condition with larger  \(\varepsilon\) and \(\mu\), since initial perturbation with much larger \(\varepsilon\) and \(\mu\) approaches more closely to the resonant condition with background \cite{Zhaoling,Lingzhao2017}.
	
	In FIG. \hyperref[fig4]{4(b)}, we show the variations of DRW as \(\Delta t\) varies when \(\varepsilon\) and \(\mu\) are fixed at different constant values.  Since the initial data with different parameters \(\varepsilon\) and \(\mu\) will yield different ${\rm GT}$, then the range of \(\Delta t\) is different for the fixed image size. According to the figure, it can be seen that when \(\varepsilon\) and \(\mu\) are fixed, DRW will decrease as a smooth curve as \(\Delta t\) increase. At the same time, if we only consider the variation of \(\varepsilon\) for fixed \(\mu\), we can see that when \(\varepsilon\) increases from $30$ to $170$, the corresponding DRW value will decrease in turn. Also, if we only consider the variation of \(\mu\) for the fixed \(\varepsilon\), the DRW value will still show a downward trend as the value of \(\mu\) increases from $0$ to  $50$. Those results are in line with the ones of FIG. \hyperref[fig4]{4(a)}.
	
	In FIG. \hyperref[fig4]{4(c)}, we show the relation between \({\rm GT}\) and \(\varepsilon\) (\(\mu\) ),  and we respectively use the exponential function with base $0.5$ and logarithmic function to fit the curves of GT about the change of \(\varepsilon\) and \(\mu\). In the Appendix B, we compare different fitting functions, and list the results to show that it is more reasonable to use the following two functions. For fixed $\mu$, we have the relation between GT and $\varepsilon$ as
	\begin{equation}\label{eq:GT(1)}
	GT=a\ln (\varepsilon)+b,
	\end{equation}
	where $a$ and  $b$ are the fitting parameters.
	For fixed $\varepsilon$, we have the relation between GT and $\mu$ as
	\begin{equation}\label{eq:GT(2)}
	GT=c\sqrt{\mu}+d,
	\end{equation}
	where $c$ and $d$ are the fitting parameters. Related fitting parameters are given in Tabel \ref{bs1} and \ref{bs2}. In Fig. \hyperref[fig4]{4(c)},  we can see that the fitting curves indeed agree well with the numerical results.

	FIG. \hyperref[fig4]{4(d)} demonstrates the variation of \(\theta\) with respect to the parameters \(\varepsilon\) or \(\mu\) respectively. Fixed the parameter $\epsilon=20$, it is shown that the angle $\theta$ almost steadily varies between $135^{\circ}$ and $140^{\circ}$. But if we fix the parameter $\mu=0.5$, it is seen that the angle $\theta$ will decrease as $\epsilon$ increases with the range approximately between $120^{\circ}$ and $135^{\circ}$. These results show the amplitude parameter $\epsilon$ will effect the angle $\theta$. Additional statistics are given in the Appendix B.
	
	\section{Conclusions}
	In this paper, we propose an automatic, fast and accurate method to measure RW pattern by deep learning. Recently, Guo et al. \cite{Guo2021} described their automated classification and positioning system for identifying localized excitations in atomic Bose-Einstein condensates by deep convolutional neural networks to eliminate the need for human image examination. They implement the detection by CNN-based image classification and least-squares fitting-based position regression. But we focus on the efficient detection of RWs on numerical solutions images of NLSEs entirely by an end-to-end deep learning framework to capture the pattern similarity of computer vision. Besides, we also get the statistics of these RW patterns based on the detection results so that we are able to predict the time interval and region angle in real cases. Our proposed term DRW can intuitively measure the the Gaussian perturbations in our experiments. Our model can be generalized for the other integrable systems with MI \cite{Manakov,Zhao2012,Chabchoub14,Baronio121318,Chen2013,Zhao2019,Tikan17,Li2013,Kartashov19,ZhangLY-21,CheFL-21,FengLT-20,MoLZ-21}.
	
	For multi-Gaussian perturbation or other weak localized forms in initial data \eqref{eq:initial} (i.e. there are two RW valleys in the images), can we still classify these rogue wave patterns and get their corresponding distribution? We leave this problem to the future work. We also hope that this work will initiate more innovative efforts in this field.
	
	\section*{Acknowlegement}
	Delu Zeng is supported by National Science Foundation of China (61571005), the Fundamental Research Program of Guangdong, China  (No. 2020B1515310023), the Science and Technology Research Program of Guangzhou, China (No. 201804010429).  Liming Ling is supported by National Natural Science Foundation of China (No. 11771151), Guangzhou Science and Technology Program(No. 201904010362).
	Li-Chen Zhao is supported by the National Natural Science Foundation of China (Contract No. 12022513, 11775176),  and the Major Basic Research Program of Natural Science of Shaanxi Province (Grant No. 2018KJXX-094).

	\section*{Appendix}
	In the Appendix, we give the outline of our proposed Peak Search method, the description of our RWD-10K dataset and the details about the setting and results of our rogue wave detection experiments.
	
	\subsection{Peak Search}
	Peak Search is our proposed fast and cluster-based algorithm to generate the rogue wave images with pseudo annotations. Through our observations, selecting points with larger modulus length from the numerical solution matrix can well filter out the crests of rogue waves.Therefore, we propose the following algorithm in Algorithm 1  to realize the goal to pre-detect the rogue waves on images. What needs to be pointed out here is that the Peak Search method only works with rogue wave matrices, which means it can not replace our RWD-Net to complete the task of detect rogue wave patterns when there are only images without any numerical solution matrices.

	
	

	\begin{algorithm}[H]
		\caption{The procedure of Peak Search algorithm.}
		\LinesNumbered
		\KwIn{The numerical solution matrix, $M=(a_{ij})$; The initical location map of the peak points, $N=(b_{ij})$ where $b_{ij}=0$ at first; }
		\KwOut{The final location map of the peak points, $N=(b_{ij})$;}
		If $a_{ij} >= \eta l$, then we set $b_{ij}=1$, where $\eta$ represents the peak factor which is a constant we specified in advance (we set $\eta =1.7$ in our experiment), and $l$ represents the level set value corresponding to the level ground plane.\\
		\label{code:fram:extract}
		For each $b_{ij}=1$, if $a_{ij}$ is the highest value point within the distance $r$, then we set $b_{ij}=2$, where $r$ denotes the comparison radius we specified in advance (we set $r =2$ in our experiment).\\
		\label{code:fram:trainbase}
		We map the numerical matrix coordinates of each peak point satisfying $b_{ij}=2$ into a weak bounding box on the image, and then zoom them in size of $20 \times 20$;
		\label{code:fram:add} \\
		\Return $N=(b_{ij})$;
	\end{algorithm}
	
	\subsection{Additional statistics}
	
\begin{figure}[h]
	\centering
	{
		\includegraphics[height=9.6cm]{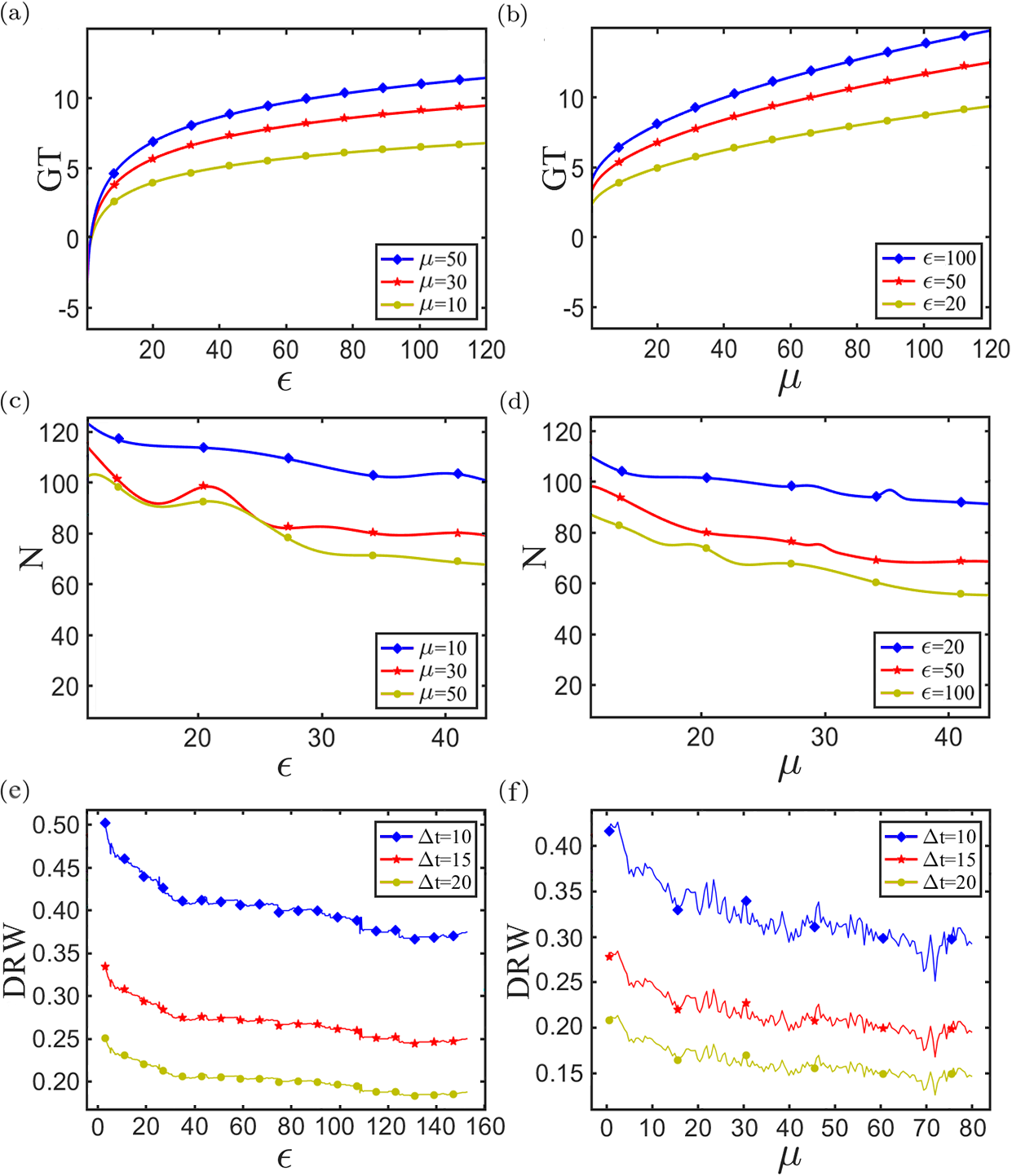}
	}
	\caption{(a-b) Variation of GT for \(\varepsilon\) or \(\mu\) when the other is fixed; (c-d) Variation of number $N$ of rogue wave in the images for \(\varepsilon\) or \(\mu\) when the other is fixed; (e) Variation of DRW for \(\varepsilon\) under different \(\Delta t\); (f) Variation of DRW for \(\mu\) under different \(\Delta t\);}\label{fig5}
	\end{figure}

	We compare different fitting functions of GT, the variation curves are shown in \hyperref[fig5]{5(a)} and the fitting parameters are shown in Table \ref{bs1} and \ref{bs2}. Besides, we give the curves showing the numbers of rogue waves in the images changing with \(\varepsilon\) and \(\mu\) in \hyperref[fig5]{5(b)}. Lastly, the variation of curves for DRW with \(\varepsilon\) and \(\mu\) are given in \hyperref[fig5]{5(c)} and \hyperref[fig5]{5(d)} under different \(\Delta t\).
	
	\begin{table}[htb]
		\caption{The parameters of fitting functions \eqref{eq:GT(1)}}
		\vspace{0pt}
		\centering
		\renewcommand\arraystretch{2}
		\setlength{\tabcolsep}{7mm}{}{
			\begin{tabular}{|c|c|c|}
				\hline
				\thead{\(\mu\)} & \thead{a}& \thead{b} \\
				\hline
				10   & 1.581   & -0.804                \\
				\hline
				30 & 2.136   & -0.766                     \\
				\hline
				50  & 2.563   &-0.819              \\
				\hline
		\end{tabular}}
		\label{bs1}
	\end{table}
	
	\begin{table}[htb]
		\caption{The parameters of fitting functions \eqref{eq:GT(2)}}
		\vspace{0pt}
		\centering
		\renewcommand\arraystretch{2}
		\setlength{\tabcolsep}{7mm}{}{
			\begin{tabular}{|c|c|c|}
				\hline
				\thead{\(\varepsilon\)} & \thead{c}& \thead{d} \\
				\hline
				20   & 0.682   & 1.900              \\
				\hline
				50 & 0.890   & 2.760              \\
				\hline
				100  &  1.042   &3.382               \\
				\hline
		\end{tabular}}
		\label{bs2}
	\end{table}

	\subsection{Experimental details}
	In this part, we show some details and results about our RWD-Net detection experiment. We use the pre-training weight file of ResNet-101 obtained from \href{https://github.com/keras-team/keras-applications/releases/tag/resnet}{https://github.com/keras-team/keras-applications/releases/tag/resnet} as the initial weight. The learning rate is set to \(1\times10^5\), the batch size is set to 4 and the epoch number is set to 20.
	Our whole training experiment runs on the machine using an NVidia TITAN RTX GPU with 11GB of GDDR6 memory.
	\begin{figure}[htb]
		\centering
		{
			\label{fig6}
			\includegraphics[height=3.3cm]{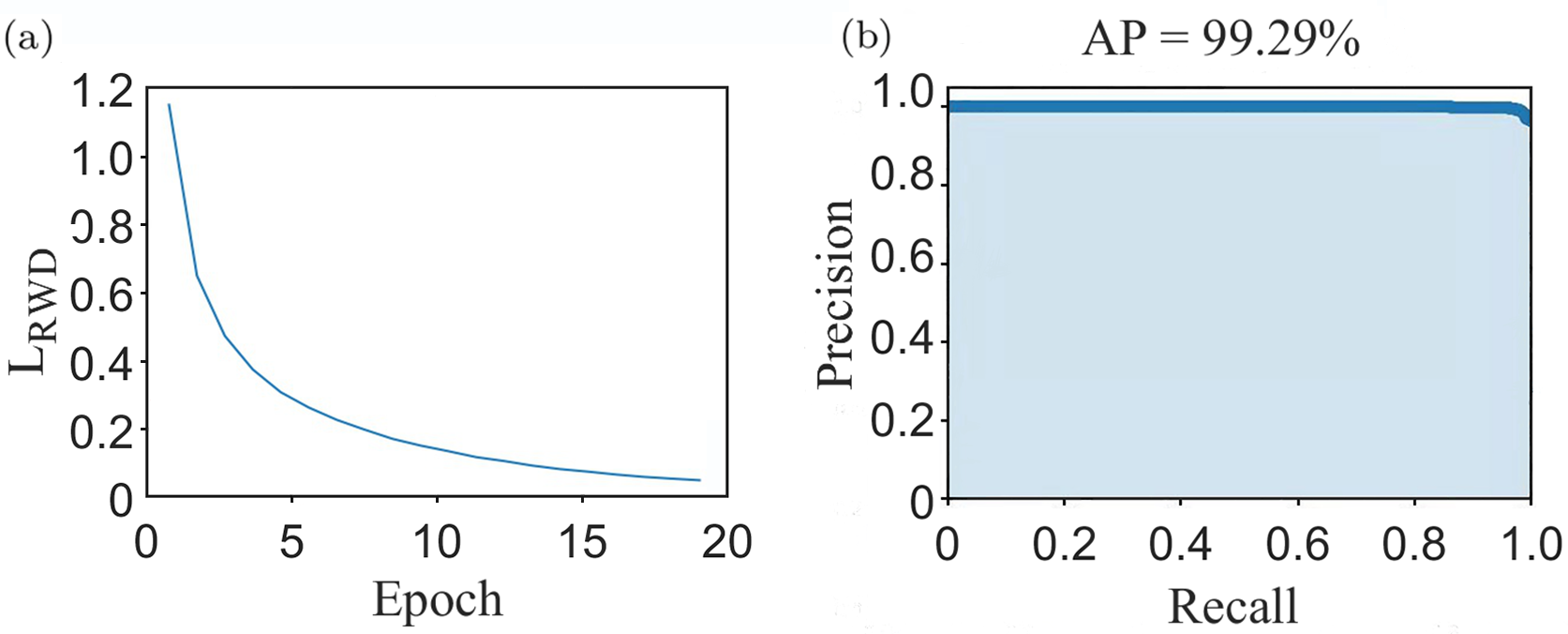}
		}
		\hspace{-0.5cm}
		\caption{(a) Evolution of the overall loss $L_{RWD}$; (b) The Precision-Recall curve of detection in test splits. }\label{fig6}
	\end{figure}
		
	We record the overall loss changes of the training set during the training process as shown in FIG. \hyperref[fig6]{6(a)}. At the end of training, the overall loss $L_{RWD}$ of the model achieves 0.071 on the train splits. Based on the trained model above, we test it on the test splits. Finally, we get 99.29\% AP at the threshold IoU set as 0.5 on the test splits and the Precision-Recall curve is shown in Fig. \hyperref[fig6]{6(b)}. 


\begin{thebibliography}{99}
	\makeatletter
	
	
	\bibitem{RW_tank}  A. Chabchoub, N. P. Hoffmann, and N. Akhmediev,
	Phys. Rev. Lett. {\bf 106}, 204502 (2011).
	
	\bibitem{prw} D. R. Solli, C. Ropers, P. Koonath, B. Jalali, Nature (London), {\bf 450}, 1054-1057 (2007).
	
	\bibitem{rw2010} B. Kibler, J. Fatome, C. Finot, G. Millot, F. Dias, G. Genty,
	N. Akhmediev, and J. M. Dudley, Nature Phys. {\bf 6}, 790 (2010).
	
	\bibitem{Dudley2014} J. M. Dudley, F. Dias, M. Erkintalo, and G. Genty,
	Nature Photon. {\bf 8}, 755 (2014).
	
	%
	%
	
	\bibitem{bec-rw} Yu. V. Bludov, V. V. Konotop, and N. Akhmediev, Phys. Rev. A, {\bf 80}, 033610 (2009);  Z. Yan, V. V. Konotop, and N. Akhmediev, Phys. Rev. A, {\bf 82}, 036610 (2010).
	
	%
	%
	%
	
	\bibitem{yanfrw} V. G. Ivancevic, Cogn. Comput. {\bf 2}, 17 (2010); Z. Yan, Commun. Theor. Phys. {\bf 54}, 947 (2010); Z. Yan, Phys. Lett. A  {\bf 375}, 4274 (2011).
	
	
	\bibitem{Kharif2009} C. Kharif  and E. Pelinovsky, Eur. J. Mech. B/Fluids, {\bf 22}, 603 (2003); P. K. Shukla, {\it et al.,} Phys. Rev. Lett. {\bf 97}, 094501 (2006); C. Kharif, E. Pelinovsky, and A. Slunyaev, {\em Rogue Waves in the Ocean} (Springer, New York, 2009); M. Onorato, {\it et al.,} Phys. Rep. {\bf 528}, 47 (2013).
	\bibitem{Onorato}
	M. Onorato, A. R. Osborne, and M. Serio,
	Phys. Rev. Lett. {\bf 96}, 014503 (2006).
	
	\bibitem{Kedziora2013} D. J. Kedziora, A. Ankiewicz, and N. Akhmediev, Phys. Rev. E {\bf 88}, 013207 (2013).
	
	\bibitem{YangY-21}B. Yang, J. Yang, Physica D: Nonlinear Phenomena, {\bf 419}, 132850 (2021).
	
	
	\bibitem{Walczak-15}P. Walczak, S. Randoux, and P. Suret, Phys. Rev. Lett. {\bf 114}, 143903 (2015).
	
	\bibitem{Gelash-18-19}A. A. Gelash, Physical Review E, {\bf 97(2)}, 022208 (2018); A. A. Gelash, D. S. Agafontsev, Physical Review E, {\bf 98(4)}, 042210 (2018); A. A. Gelash, D. gafontsev, V. Zakharov, et al., Physical Review papers, {\bf 123(23)}, 234102 (2019).
	
	\bibitem{Soto-Crespo-16}J. M. Soto-Crespo, N. Devine, N. Akhmediev, Phys. Rev. Lett. {\bf 116(10)}, 103901 (2016).
	
	
	\bibitem{MI-1967} T. B. Benjamin, Proc. R.  Soc. A {\bf 299}, 59-75 (1967);  T. B. Benjamin, and J. E. Feir, J. Fluid  Mech. {\bf 27}, 417-430 (1967).
	
	\bibitem{Baronio} F. Baronio, M. Conforti, A. Degasperis, S. Lombardo, M. Onorato, and S. Wabnitz,
	Phys. Rev. Lett. {\bf 113}, 034101 (2014).
	
	
	\bibitem{MI-nonlinear-MI}
	V. E. Zakharov and A. A. Gelash,
	Phys. Rev. Lett. {\bf 111}, 054101 (2013);
	G. Biondini, and D. Mantzavinos,
	Phys. Rev. Lett. {\bf 116}, 043902 (2016).
	
	\bibitem{Jordan-15}M. I. Jordan, T. M. Mitchell, {\bf 349(6245)}, 255-260 (2015).
	
	\bibitem{Narhi-18}M. N\"arhi, L. Salmela, J. Toivonen, C. Billet, J.M. Dudley, and G. Genty, Nature communications, {\bf 9(1)}, 1-11 (2018).
	
	
	\bibitem{Amil-19}P. Amil, M. C. Soriano, and C. Masoller, Chaos: An Interdisciplinary Journal of Nonlinear Science, {\bf 29(11)}, 113111 (2019).
	
	\bibitem{Qi-20}D. Qi, A. J. Majda, Proceedings of the National Academy of Sciences, {\bf 117(1)}, 52-59 (2020).
	
	\bibitem{Mohamad-18}M. A. Mohamad, T. P. Sapsis, Proceedings of the National Academy of Sciences, {\bf 115(44)}, 11138-11143 (2018).
	
	\bibitem{Dematteis-18}G. Dematteis, T. Grafke, E. Vanden-Eijnden, Proceedings of the National Academy of Sciences, {\bf 115(5)}, 855-860 (2018).
	
	\bibitem{Majda-19}A. J. Majda, M. N. J. Moore, D. Qi, Proceedings of the National Academy of Sciences, {\bf 116(10)}, 3982-3987 (2019).
	
	\bibitem{Salmela-20}L. Salmela, C. Lapre, J. M. Dudley, G. Genty, Scientific Reports, {\bf 10(1)}, 1-8 (2020).
	
	\bibitem{Tao}J. X. Tao, J. B. Huang, L. Yu, Z. K. Li, H. S. Liu, B. Yuan, D. L. Zeng, Food Hydrocolloids, {\bf 74(28)}, 151-158 (2017).
	
	\bibitem{WangLZF-21}R. Q. Wang, L. M. Ling, D. L. Zeng and B. F. Feng, Communications in Nonlinear Science and Numerical Simulation, {\bf 101}, 105896 (2021).
	
	\bibitem{Zhaoling} L. C. Zhao and L. M. Ling, J. Opt. Soc. Am. B   \textbf{33}, 850-856 (2016).
	
	\bibitem{GaoZYLY-20} P. Gao, L. C. Zhao, Z. Y. Yang, X. H. Li, and W. L. Yang, Opt. Lett. {\bf 45}, 2399-2402 (2020)
	
	
	\bibitem{Yang-10}J. Yang, Nonlinear waves in integrable and nonintegrable systems. Society for Industrial and Applied Mathematics (2010).
	
	\bibitem{Peregrine1983} D. Peregrine, J. Aust. Math. Soc. B, Appl. Math. {\bf 25}, 16 (1983).
	
	
	\bibitem{Akhmediev2009a}N. Akhmediev, A. Ankiewicz, and J. Soto-Crespo,
	Phys. Rev. E {\bf 80}, 026601 (2009).
	
	
	\bibitem{Biondini-18}G. Biondini, X. Luo, Phys. Lett. A {\bf 382(37)}, 2632-2637 (2018).
	
	
	\bibitem{He2016}K. He, X. Zhang, S. Ren, J. Sun and Microsoft Research, 2016 IEEE Conference on Computer Vision and Pattern Recognition, 770-778 (2016)
	
	\bibitem{He2017b}T. Lin, P. Dollar, R. Girshick, K. He, B. Hariharan and S. Belongie, 2017 IEEE Conference on Computer Vision and Pattern Recognition, 936-944 (2017).
	
	\bibitem{He2017a}T. Lin, P. Goyal, R. Girshick,K. He and P. Dollar, 2017 IEEE Conference on Computer Vision and Pattern Recognition, 2999-3007 (2017).
	
	\bibitem{Goldman2019}E. Goldman, R. Herzig, A. Eisenschtat, O. Ratzon, I. Levi, j. Goldberger and T. Hassner, 2019 IEEE Conference on Computer Vision and Pattern Recognition, 5227-5236 (2019).
	
	\bibitem{smoothL1}R. Girshick, 2015 IEEE international conference on computer vision, 1440-1448 (2015).
	
	\bibitem{He2015}S. Ren, K. He, R. Girshick, et al., IEEE Transactions on Pattern Analysis and Machine Intelligence, {\bf 39(6)}, 1137-1149 (2017).
	
	\bibitem{gelatinization}J. Tao, J. Huang, L. Yu, Z. Li, H. Liu, B. Yuan, et al., Food Hydrocolloids, {\bf 74}, 151-158 (2017).
	
	\bibitem{Lingzhao2017} L. M. Ling, L. C. Zhao, Z. Y. Yang, B. Guo, Phys. Rev. E {\bf 96}, 022211 (2017).
	
	\bibitem{Guo2021}S. Guo, A. R. Fritsch, C. Greenberg, et al., Machine Learning: Science and Technology, {\bf 2}, 035020 (2021).
	
	\bibitem{Manakov} S. V. Manakov, Zh. Eksp. Teor. Fiz. {\bf 67}, 543 (1974) [Sov. Phys. JETP, {\bf 38}, 248 (1974)].
	
	\bibitem{Zhao2012} B. L. Guo and L. M. Ling, Chin. Phys. Lett. {\bf 28}, 110202 (2011); L. C. Zhao and J. Liu, J. Opt. Soc. Am. B {\bf 29},
	3119 (2012); L. C. Zhao and J. Liu, Phys. Rev. E {\bf 87}, 013201 (2013).
	
	\bibitem{Chabchoub14} A. Chabchoub and M. Fink, Phys. Rev. Lett. {\bf 112}, 124101 (2014); A. Przadka, S. Feat, P. Petitjeans, V. Pagneux, A. Maurel, and M. Fink, Phys. Rev. Lett. {\bf 109}, 064501 (2012).
	
	\bibitem{Baronio121318} F. Baronio, A. Degasperis, M. Conforti, and S. Wabnitz, Phys. Rev. Lett. {\bf 109}, 044102 (2012); F. Baronio, M. Conforti, A. Degasperis, and S. Lombardo, Phys. Rev. Lett. {\bf 111}, 114101 (2013); S. Chen, Y. Ye, J. M. Soto-Crespo, Ph. Grelu, and F. Baronio, Phys. Rev. Lett. {\bf 121}, 104101 (2018).
	
	\bibitem{Chen2013} S. Chen and L.Y. Song, Phys. Rev. E {\bf 87}, 032910 (2013); S. Chen, Phys. Lett. A {\bf 378}, 2851 (2014); S. Chen and D. Mihalache, J. Phys. A {\bf 48}, 215202 (2015); S. Chen, X. M. Cai, P. Grelu, J. Soto-Crespo, S. Wabnitz, and F. Baronio, Opt. Express {\bf 24}, 5886 (2016).
	\bibitem{Zhao2019}  L. C. Zhao, L. Duan, P. Gao, Z. Y. Yang, EuroPhys. Lett. {\bf 125}, 40003 (2019).	
	\bibitem{Tikan17} A. Tikan, C. Billet, G. El, A. Tovbis, M. Bertola, T. Sylvestre, F. Gustave, S. Randoux, G. Genty, P. Suret, and J. M. Dudley, Phys. Rev. Lett. {\bf 119}, 033901 (2017).
	
	\bibitem{Li2013} L. Li, Z. Wu, L. Wang, and J. He, Ann. Phys. {\bf 334}, 198 (2013); J. He, H. Zhang, L. Wang, K. Porsezian, and A. Fokas, Phys. Rev. E {\bf 87}, 052914 (2013).
	\bibitem{Kartashov19} Y. V. Kartashov, V. V. Konotop, M. Modugno, and E. Ya. Sherman, Phys. Rev. Lett. {\bf 122}, 064101 (2019).
	
	\bibitem{ZhangLY-21}G. Q. Zhang, L.M. Ling, Z.Y. Yan, Journal of Nonlinear Science, {\bf 31(5)}, 1-52 (2021).
	
	\bibitem{CheFL-21}Y. R. Chen, B. F. Feng, L. M. Ling, Physica D: Nonlinear Phenomena, {\bf 424}, 132954 (2021).	
	
	\bibitem{FengLT-20}B. F. Feng, L. M. Ling, D. A. Takahashi, Studies in Applied Mathematics, {\bf 144(1)}, 46-101 (2020).
	
	\bibitem{MoLZ-21} Y. F. Mo, L. M. Ling and D. L. Zeng, Phys. Lett. A, 127739 (2021).
	
	
\end{thebibliography}
\end{document}